\documentclass[a4paper, 10pt, conference]{ieeeconf}      

\IEEEoverridecommandlockouts                              

\usepackage{graphics} 
\usepackage{graphicx}
\usepackage{amsmath} 
\usepackage{amssymb}  

\title{\LARGE \bf
Instance-specific 6-DoF Object Pose Estimation \protect\\ from Minimal Annotations
}

\author{Rohan P. Singh$^{1,2}$, Iori Kumagai$^{1}$, Antonio Gabas$^{2,3}$ \protect\\ Mehdi Benallegue$^{1}$, Yusuke Yoshiyasu$^{3}$, Fumio Kanehiro$^{1,2}$
\thanks{$^{1}$Rohan P. Singh, Iori Kumagai, Mehdi Benallegue and Fumio Kanehiro are with HRG (Humanoid Research Group), National Institute of Advanced Industrial Science and Technology (AIST), Japan
        {\tt\small rohan-singh@aist.go.jp}}%
\thanks{$^{2}$Rohan P. Singh, Antonio Gabas and Fumio Kanehiro are with University of Tsukuba, Ibaraki, Japan}%
\thanks{$^{3}$Antonio Gabas and Yusuke Yoshiyasu are with CNRS-AIST JRL (Joint Robotics Laboratory) UMI3218/RL, National Institute of Advanced Industrial Science and Technology (AIST), Japan}%
}

\begin{document}

\maketitle
\thispagestyle{empty}
\pagestyle{empty}

\begin{abstract}

In many robotic applications, the environment setting in which the 6-DoF pose estimation of a known, rigid object and its subsequent grasping is to be performed, remains nearly unchanging and might even be known to the robot in advance. In this paper, we refer to this problem as instance-specific pose estimation: the robot is expected to estimate the pose with a high degree of accuracy in only a limited set of familiar scenarios. Minor changes in the scene, including variations in lighting conditions and background appearance, are acceptable but drastic alterations are not anticipated. To this end, we present a method to rapidly train and deploy a pipeline for estimating the continuous 6-DoF pose of an object from a single RGB image. The key idea is to leverage known camera poses and rigid body geometry to partially automate the generation of a large labeled dataset. The dataset, along with sufficient domain randomization, is then used to supervise the training of deep neural networks for predicting semantic keypoints. Experimentally, we demonstrate the convenience and effectiveness of our proposed method to accurately estimate object pose requiring only a very small amount of manual annotation for training.

\end{abstract}

\section{INTRODUCTION}

For a robot to grasp and manipulate any object in its surrounding environment, it is essential for it to estimate the position and orientation of the object relative to itself - often through the use of its vision sensors. In recent years, advances in deep learning based approaches have used powerful convolutional neural networks (CNNs) to process the input image data and generate a pose prediction \cite{pavlakos17object3d, peng2019pvnet, xiang2018posecnn}. The networks attempt to learn a mapping from the high-dimensional input feature space to an output space where the learning process is generally needed to be supervised through a set of labeled samples. To avoid overfitting of the network on the training dataset and achieve a good generalization, the domain space needs to be sufficiently sampled meaning that the labeled training data points should consist of enough input feature variations. In the case of CNNs designed for detecting pre-defined object keypoints in the input RGB image \cite{rhodin2018learning, newell2016stacked, xiang2018posecnn, peng2019pvnet, nibali20193d} this implies that the training dataset should be composed of annotated images of the object in many different backgrounds with varying lighting conditions and surrounding environments. Hence, the required training dataset quickly grows huge in size. A central problem, then, is the generation or accumulation of this labeled dataset with minimal manual effort and time consumed \cite{rhodin2018learning, kocabas2019epipolar, Besl:1992:MRS:132013.132022}.\par

\begin{figure}[t]
  \includegraphics[width=\linewidth]{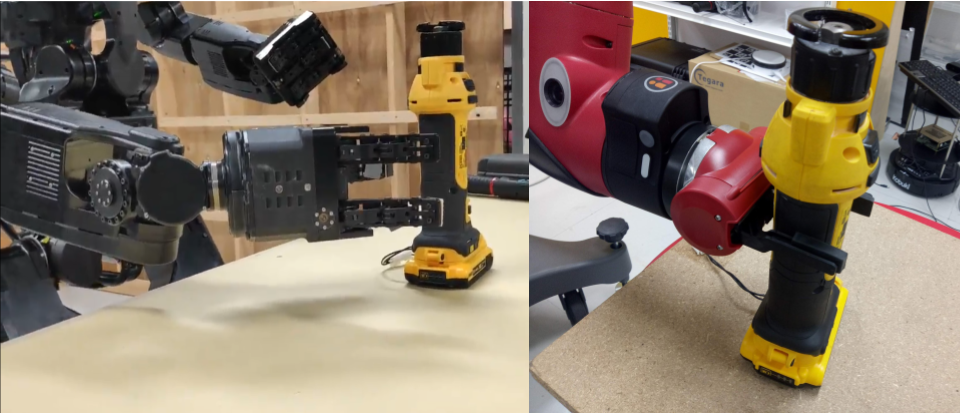}
  \caption{Grasping of a DEWALT cutout tool with known 3D model by (a) HRP-5P humanoid on the left and (b) Sawyer arm on the right.}
  \label{figure:grasping1}
\end{figure}

Further, in applications such as fetch and deliver robots in household environments, pick-and-place of objects by robot arms in industrial environments, grasping tasks by autonomous cooking robots, the object under focus generally has a known 3D model. Moreover, the environment in which the pose estimation task is to be performed is also, more or less, known and is not expected to change drastically over time. The background scene is nearly fixed. In other words, for such applications, we argue not only that there is a strong overlap between the training and the testing domains but also that this overlapping domain lies in a familiar, narrow region; and so, the network needs be trained to make accurate predictions in this region. Hence, in these case, the demanded variation in the training data is significantly lowered as the focus is on instance-specific pose estimation of a known object in a limited number of scenarios. By instance-specific, we mean that the object and its surrounding environment in the image view is expected to vary in a very narrow patch in the domain space. \par
In our work, the objective is to generate training data for this case by presenting a very convenient, semi-automated labeled dataset generation technique to ultimately train a object pose estimation pipeline in a known, local environment. Our method uses known camera poses and rigid-body geometry to triangulate pre-defined model keypoints in a few images, and then reprojecting them onto other images. We gather data for a set of varying backgrounds and lighting conditions and utilize domain randomization for further generalization. For the pose estimation task, as demonstrated with remarkable performance in many recently proposed works, we too employ a multi-stage semantic keypoint based approach. We show that the generated training dataset can be used to effectively train the two networks of the framework - a 2D object detector \cite{redmon2016you} followed by a ``stacked-hourglass" keypoint detector \cite{newell2016stacked} - to recover the 6-DoF pose of an object from a single RGB image at near real-time speeds with robustness to limited occlusions and background variations.\par
Our contributions through this research, are thus summarized as follows:
\begin{itemize}
\item We present a novel method of automatically generating huge pose annotated datasets from minimal manual annotation of unlabeled RGB images. The labeled dataset comprises of real images captured in the known, local environment as well as samples created through domain randomization of background.
\item We propose improvements upon the standard semantic-keypoint based approach for 6-DoF pose estimation \cite{pavlakos17object3d} in situations when the point cloud is available, by recovering pose from 3D positions of the detected 2D keypoints using Orthogonal Procrustes analysis and a spectral-clustering based outlier rejection scheme.
\item We verify experimentally the practicality and quick deployability of our framework for robotic grasping tasks - using the HRP-5P humanoid robot to grasp a hand-operated tool for our experiments (example grasps shown in Fig. 1).

\end{itemize}

\section{RELATED WORK}
\noindent
\textbf{6-DoF pose estimation.} The problem of pose estimation of hand-operated tools, appliances and everyday objects has drawn significant research interest. If the sensory data comprises of depth information (such as point cloud data) and the object model is known, methods such as Iterative Closest Point (ICP) \cite{Besl:1992:MRS:132013.132022} try to align the model in the 3D scene point cloud by minimizing a distance function. Algorithms based on ICP have demonstrated sufficient accuracy for grasping tasks but their slow convergence speeds and susceptibility to local minima make them unfavourable for many applications. Hence most methods either follow a holistic template-based approach or, more recently, a feature-based approach.\par
Template-based methods use matching schemes and a set of template images of the object generated by rendering its 3D model from various viewpoints. At runtime, they attempt to compare the input image with the templates and obtain the best match by computing similarity scores. Holistic deep learning methods have also been proposed which aim to estimate the position and orientation of the object in a single-shot. While such methods have achieved high accuracy, their performance remains poor when the object is partially occluded. On the other hand, in feature-based methods \cite{pavlakos17object3d, rad2017bb8, tekin2018real}, the pose estimation task is broken into two-stages: first detecting 2D features or keypoints on the object as viewed in the image and then using the object 3D model to establish a set of 2D-to-3D point correspondences. Recovering the full pose of the object in the camera frame is then formulated as a Perspective-n-Point problem, which has already been studied extensively. As there have been significant advancements in neural network architectures for 2D semantic keypoint predictions in RGB images, state-of-the-art techniques are largely built upon such a pipeline. For instance PoseCNN proposed by Xiang et. al. \cite{xiang2018posecnn} extracts the 2D center of the object by voting for the center using a vector field and regressing to a quaternion for estimating the rotation. The more recent PVNet method \cite{peng2019pvnet} regresses to vector fields for each of the pre-defined object keypoints and uses these vectors to vote for keypoint locations using RANSAC. It then uses an uncertainty-driven P\textit{n}P to find the full pose. \par

\begin{figure}[t]
  \includegraphics[width=\linewidth]{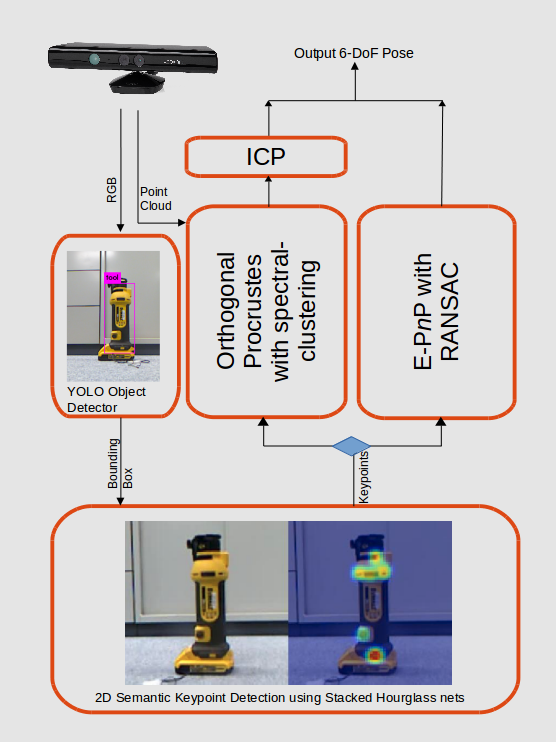}
  \caption{Overview of proposed pose estimation pipeline. Our method can compute the 6-DoF pose either by solving a P\textit{n}P problem on the 2D keypoints or through the Orthogonal Procrustes analysis on 2 sets of 3D points. The latter approach is possible only when the point cloud is available. Both approaches have different outlier rejection mechanisms.}
  \label{figure:approach1}
\end{figure}

Pavlakos et. al. \cite{pavlakos17object3d} propose to use a stacked-hourglass network which predicts heatmaps for semantic keypoint on the object and then fit a deformable shape-model to the 2D detections. In this paper, we adopt a framework similar to theirs for 2D semantic keypoint prediction. However, we differ from \cite{pavlakos17object3d} in two ways. First, as we focus on known instances of objects, we can directly solve a P\textit{n}P problem on the 2D keypoints to obtain the pose. Second, for cases when the scene point cloud is available during test time, we propose to optionally use a 3D keypoint based approach to find the pose. \par
\noindent
\textbf{Generating annotations.} A major drawback of the keypoint based state-of-the-art techniques mentioned above is their need for tedious manual annotations. Most methods mentioned thus far require supervision through 6-DoF pose-labeled data. Techniques have been proposed for reducing the dependence on annotated data \cite{rhodin2018learning} or generating 3D ground truth data using 2D pose estimators \cite{pavlakos17harvesting, kocabas2019epipolar}. Nevertheless, these methods still require either 2D ground truth data or a small amount of 3D data, which is difficult to obtain. For instance, \cite{pavlakos17object3d} generate the 2D keypoint labels by careful manual 3D model to point cloud alignment. To generate the bounding box and pose labels in a semi-automated way Suzui et. al. in \cite{suzui2019toward} also propose a technique using pose markers. However, contrary to their technique, our method does not require known transformation between the markers and the object, which greatly simplifies the process.

\section{APPROACH}

In this section, we first describe the approach adopted by us for localizing 2D semantic keypoints in the RGB image (in Section III-A) and for subsequently recovering the 6-DoF pose (in Section III-B). To overcome the challenge of manual annotation of 2D keypoints, we also present our novel method of semi-automating the procedure (in Section III-C). An overview of the proposed architecture is shown in Fig. 2.

\begin{figure}[t]
  \includegraphics[width=\linewidth]{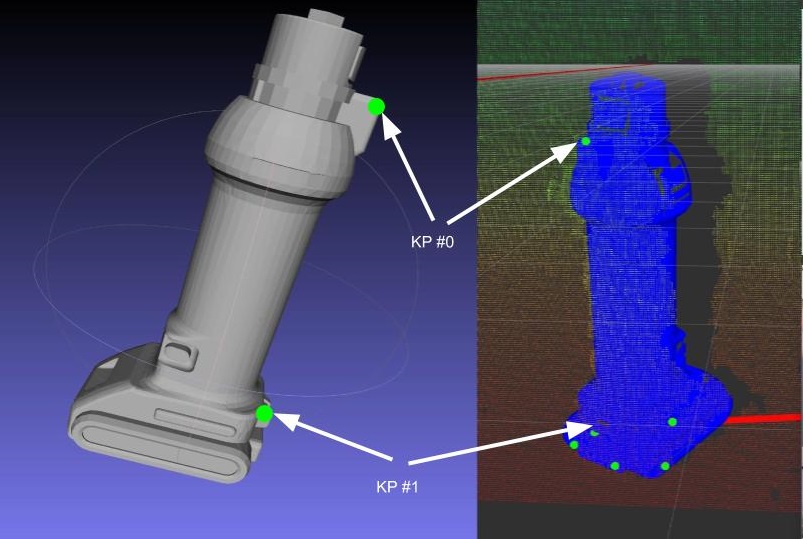}
  \caption{Selection of keypoints on the DEWALT cutout tool and corresponding 3D point estimated using 2D detection results and point cloud data.}
  \label{figure:keypoints}
\end{figure}

\subsection{Semantic keypoint localization}

Inspired by the success of the “stacked-hourglass” deep neural network architecture in human-pose estimation \cite{newell2016stacked, nibali20193d, kocabas2019epipolar}, and also more recently in generic 3D object pose estimation \cite{pavlakos17object3d}, we adopt the same network design for keypoint prediction. In our pipeline, the semantic keypoint prediction network consists of two hourglass modules stacked end-to-end where the output of the last hourglass module is a set of “heatmaps”. Here each pixel of every heatmap holds the probability of the existence of the corresponding keypoint at that pixel location. During the testing phase, the most likely position of the 2D keypoint is taken to be the coordinates of the peak in the corresponding heatmap.\par
The input to the hourglass network is an RGB image of equal width and height cropped to tightly fit the object of interest in the image frame. For this, any of the existing highly accurate 2D object detectors can be trained and used to get the bounding box around the object, e.g. SSD \cite{liu2016ssd}, YOLO \cite{redmon2016you}, Faster-RCNN \cite{ren2015faster}, etc. In this paper, we integrate the YOLO object detector in the ROS pipeline which forwards the bounding box coordinates to the succeeding hourglass network, which eventually makes the heatmap predictions. Hence, the output of this stage is a 2D keypoint pose configuration of the object.\\\\
\textbf{Keypoint selection.} While defining the keypoint on the 3D model, we need to ensure that the keypoints are evenly distributed on all faces of the object. A good spread ensures stable pose estimation for both approaches - through Procrustes Analysis on 3D points or through the P\textit{n}P algorithm on 2D points. Experimentally, we find that for our method, choosing a set of 20 evenly spread keypoints on the model are required for providing sufficient robustness. Fig. 3 (left) shows the keypoints selected on the 3D model.

\subsection{Pose Estimation with Outlier Rejection}

Once the 2D keypoints are localized in the input image, the pipeline branches into two possible routes:\\\\
\textbf{P\textit{n}P approach based on 2D keypoints.} In scenarios where the scene point cloud data is unavailable or there is significant occlusion due to scene clutter, a straight-forward way to compute the 6-DoF pose is to use an off-the-shelf P\textit{n}P algorithm such as EP\textit{n}P \cite{lepetit2009epnp} on the 2D keypoints.  Outlier rejection in this case can be achieved through RANSAC \cite{fischler1981random}. This approach enables good pose estimation even in the presence of minor occlusions of some keypoints, nevertheless subject to the condition that the keypoints are correctly localized in the previous step. In our implementation, we use the OpenCV solvePnPRansac() function relying on the EP\textit{n}P method proposed by Lepetit et. al in \cite{lepetit2009epnp}.\\\\
\textbf{Orthogonal Procrustes analysis based on 3D keypoints.} In our experiments, however, we observed that the RANSAC scheme can occasionally fail to adequately reject the outliers in which case, the computed pose through P\textit{n}P might have very large errors that are unable to be corrected through ICP. Hence, in this paper, we propose an alternative approach to obtain the 3D pose from 2D keypoints which can be used if the point cloud data of the scene is available and if the scene is largely uncluttered (such that there is no obstruction blocking the path from the camera optical center to the 3D keypoint). We use the 2D keypoint locations along with the camera intrinsics and the point cloud to find the 3D position of the keypoints in the camera frame. Given a 2D keypoint $\textbf{k} = (k_x, k_y)^T$, we first extract its 3D position $(T_x, T_y, T_z)$ as follows:

$$
T_{x} = \frac{T_{z}}{f_{x}}\left ( c_{x} - p_{x}\right ) \eqno{(1)}
$$
$$
T_{y} = \frac{T_{z}}{f_{y}}\left ( c_{y} - p_{y}\right ) \eqno{(2)}
$$

where $(p_{x}, p_{y})^T$ is the principal point and $f_{x}$ and $f_{y}$ are the local lengths of the camera. $T_{z}$ is the depth of the keypoint in 3D space estimated by projecting the 2D keypoint into the 3D space and finding its intersection with the object point cloud.\par

\begin{figure*}
  \includegraphics[width=\textwidth]{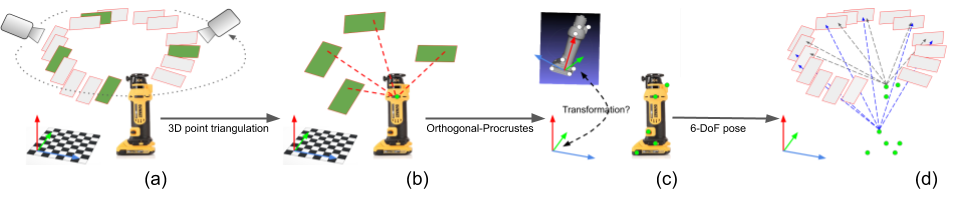}
  \caption{Overview of the semi-automated label generation method. The process begins at (a) where we have a fixed position of the object relative to the marker and the camera captures the RGB images from various viewpoints. Several frames with distinct viewpoints are randomly selected for manual annotation (shown in green). (b) Marked 2D keypoints are used to triangulate the point in 3D space (using Eq. 3). This is done for at least 4 points. (c) 3D pose of the object in marker frame is recovered through Orthogonal Procrustes analysis. (d) Rest of the model keypoints are projected to all of the originally captured frames. Note that here we needed to annotate a total of only 20 keypoints (5 each in 4 images) to essentially obtain 100s of annotated images. This process is repeated for several different backgrounds and lighting conditions.}
  \label{figure:approach2}
\end{figure*}

Next, the pipeline employs an outlier rejection scheme based on spectral clustering proposed in \cite{bandari2004efficient} to remove the erroneously estimated 3D points. Finally, the full pose is analytically obtained from the inliers through orthogonal Procrustes analysis. Here we use the popular Horn's method \cite{horn1987closed} to compute the rigid transformation between the observed set of 3D keypoints and the defined keypoints on the object model. Empirically, we show that in certain circumstances, this approach can provide more reliability to the pipeline, in comparison to a standard P\textit{n}P based approach.\par
As an optional final stage, we can use an Iterative Closest Point (ICP) algorithm to achieve finer results but at a slower frame rate \cite{Besl:1992:MRS:132013.132022}.

\subsection{Semi-automated dataset generation}

We introduce a convenient process to generate thousands of labeled images from just a few manual annotations (as illustrated in Fig. 4). To train the two networks used in our pipeline we need to create a dataset consisting of the RGB images and the associated labels for the object keypoint coordinates as well as the bounding box coordinates.\par
To this end, we first capture a set of multiple raw videos of the objects, placed in an environment similar to the test environment, by moving the camera in the scene around the object so as to capture different views of the object. We deduce the camera poses by placing a checkerboard marker anywhere in the scene and it is not required to know where exactly is the marker placed in the scene. The only condition we must ensure, however, is that the relative pose transformation between the marker and the object must remain constant throughout the video. Thus, we capture several videos in different settings - changing the orientation of the object, moving to a new background and different lighting conditions. We consider the marker coordinate frame as the global frame all through our approach and all geometrical measurements are done in this frame.\par
For a few selected image frames in each video, we triangulate the 3D positions of at least 5-7 object keypoints by using the known camera poses and camera intrinsic parameters. The triangulation is done by manually annotating only the visible keypoints in the selected images and finding the closest point of intersection of the rays emanating from the camera optical centers passing through keypoint pixel coordinates in the virtual image planes into the direction of keypoints actual position in the 3D space. Due to small errors in manual annotations, camera poses and intrinsics, the rays corresponding to each keypoint do not perfectly intersect at one single point; so instead the mutually closest point is found by solving a least-squares problem. More formally, if the camera centers are denoted by $ T\in \mathbb{R}^{3 \times N_{c}} $ and the unit vectors along each of the non-intersecting rays for the k-th keypoint and l-th image frame by $ \hat{v}_{k,l}$, then the closest point of intersection $Q_{k} \in \mathbb{R}^3$ can be given by:

$$
Q_{k}= \left ( \sum_{l}^{N_{c}} \left ( I - \hat{v}_{k,l} \hat{v}_{k,l}^{T} \right )\right )^{-1} \left ( \sum_{l}^{N_{c}} \left ( I - \hat{v}_{k,l} \hat{v}_{k,l}^{T} \right )T_{l}\right ) \eqno{(3)}
$$

where, $ l\in\{1,...N_{c}\} $ and $ k\in\{1,...N_{k}\} $. $N_{c}$  is the number of selected images and $N_{k}$ is the number of keypoints manually annotated keypoints in each image. In our experiments, we set $N_{c}$  = 5 and $N_{k}$ = 6 (see Fig. 4).\par
Next, as the object has remained fixed relative to the marker in all frames of the video, Procrustes analysis is used to find the 6-DoF object pose using the triangulated points and the object model information. Once we obtain the 3D pose of the object in the marker frame in the video sequence, we can project all the object’s keypoints to all image frames extracted from the video. In this way, we obtain annotations for all the keypoints defined on the object model, which is useful as the network can be trained to predict even the occluded, non-visible keypoints.\par
Labels for bounding boxes can be generated from here as follows: using the object pose obtained above, project all 3D model vertices on the corresponding image. Taking the minimum and maximum of the projected points in x- and y-axis, we get the edges of a tightly bound box around the object.\\\\
\textbf{Domain Randomization.} To bridge the gap between the train dataset so generated and the actual test case data, we use a recently proposed Domain Randomization (DR) method \cite{tobin2017domain}. To create the domain randomized set, we first segment out the foreground object from several images-pose pairs. This is done by projecting the 3D object model on the 2D image and then cropping the image inside a convex hull of the projected model vertices. The segmented object images are then pasted on a subset of 2000 images from the COCO dataset \cite{lin2014microsoft}, at varying orientations and color intensities. Ultimately, for each object the final training dataset comprises of 2 subsets: real images from the video sequences captured in the test environment and the domain randomized images, which as we explain in the next section, can be combined in different proportions to effectively train the network.

\section{EXPERIMENTAL RESULTS}
For evaluating our method's effectiveness and its applicability for robotic grasping tasks in familiar environments, we concentrated on the task of grasping and picking-up of a DEWALT cutout tool by the HRP-5P humanoid robot. The environment was set up to represent a construction site and the object was placed on the top of a gypsum board.

\subsection{Training datasets}
For training, we recorded a total of 3 videos around the object under focus along with other stray objects and a checkerboard marker placed in different orientations and lighting conditions. As our method requires, the position and orientation of the focused object relative to the marker was kept constant through each individual video, while the stray objects were freely moved randomly to different locations. Next, from each video: we extracted 300 frames on average and chose 5 frames for the manual annotation process - visually locating and labelling at least 6 keypoints in each frame from a total of 20 pre-defined keypoints on both the objects. The frames for manual annotation were needed to be chosen carefully, as we must ensure that the viewpoints are distinct but still at least 6 common keypoints are visible in all frames.\par
Our proposed semi-automated annotation generation technique, then, yielded labels for rest of the nearly 900 images from the recorded videos and additionally, 2000 domain randomized labeled samples. We experimented with different ratios of the domain randomized samples to the real environment samples to compile a whole dataset of size 2000 samples. Example images from the train dataset thus generated are shown in Fig. 5. \par
Two validation sets were created by taking 200 real-world images in different backgrounds (referred to as BG1 and BG2) and annotating them manually. The average and peak errors on the validation sets were calculated in terms of position and RPY representation of orientation (tabulated in Table I and Table II).

\begin{figure}[t]
  \includegraphics[width=\linewidth]{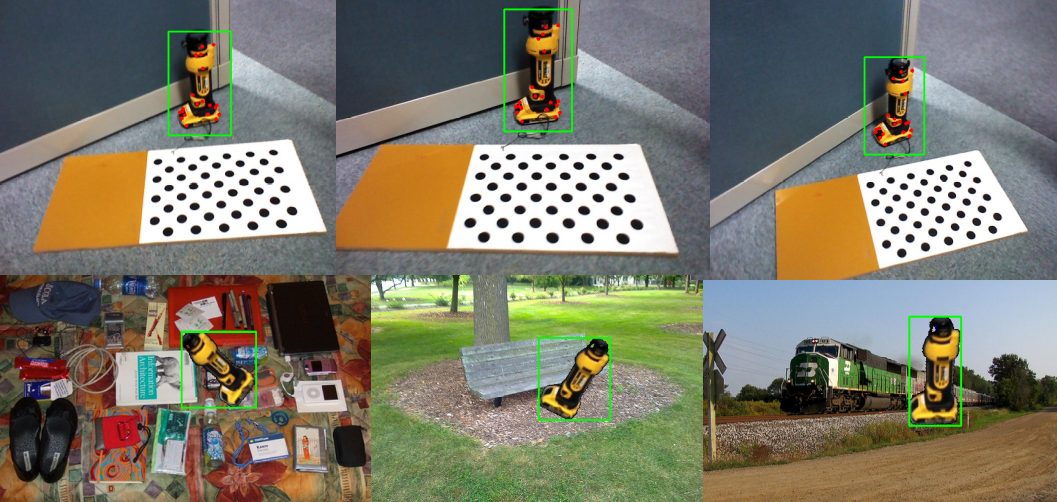}
  \caption{Example images from dataset generated through our proposed method. The top row shows keypoint labeled images captured in the test scenerio. The bottom row shows examples of DR images.}
  \label{figure:app4s}
\end{figure}

\subsection{Robotic Grasping}
To validate the accuracy of the estimated pose using our pipeline, we perform grasping and picking up of the objects using the HRP-5P humanoid robot. We performed 10 grasping experiments for the same object in different orientations using the Horn's approach with 3D keypoints in an uncluttered scene. We succeeded in 8 out of 10 trials. However, due to the robot's limitations in grasping approaches and planning some orientations of the object could not be tested. We observed that when the object is standing upright, the grasping attempts generally succeed owing to an accurate pose estimation. On the Saywer platform grasping attempts succeeded with more ease (4 out of 4 trials) - as a consequence of the wide grippers.

\begin{table}[h]
\caption{AVERAGE ERRORS IN POSE ESTIMATION OF DEWALT TOOL}
\label{table_grasping_results_I}
\begin{center}
\begin{tabular}{c | c c}
\hline
& Position & Orientation\\
& (in \textit{meters}) & (in \textit{degrees})\\
\hline
BG1 & 0.03 & -2.275, -3.266, 0.944\\
BG2 & 0.027 & -3.357, 9.561, 1.53\\
\hline
\end{tabular}
\end{center}
\end{table}

\begin{table}[h]
\caption{PEAK ERRORS IN POSE ESTIMATION OF DEWALT TOOL}
\label{table_grasping_results_II}
\begin{center}
\begin{tabular}{c | c c}
\hline
& Position & Orientation\\
& (in \textit{meters}) & (in \textit{degrees})\\
\hline
BG1 & 0.036 & 8.723, 10.364, 5.206\\
BG2 & 0.033 & 7.084, 11.057, 3.437\\
\hline
\end{tabular}
\end{center}
\end{table}

\section{CONCLUSION}
In this paper, we have demonstrated a very practical, highly accurate and near real-time ROS pipeline for estimating the 6-DoF pose of known, rigid objects. Further, we presented a very efficient and practical technique of generating huge annotated datasets from very little manual effort. Experimentally, we showed that the dataset generated from the partially automated process can be effectively used to train the pipeline to a high degree of accuracy, which is sufficient to perform robot grasping and manipulation tasks. We also showed that an alternative approach of recovering the 6-DoF pose from 2D keypoints using point cloud data and a spectral clustering based outlier rejection mechanism provides added robustness to the system.\par
As our current system was implemented on ROS, it can be easily integrated with any ROS based robotic framework. We tested the grasping of a drywall-cutout tool - by the HRP-5P humanoid robot and the Sawyer robotic arm and achieved impressive rates of successful grasping attempts.\par
We believe a future step in this direction would be to introduce photorealistic synthetic data to the training process. This will further improve the performance of the system in extreme lighting conditions and rare poses.





\bibliographystyle{IEEEtran}
\bibliography{IEEEabrv,bibliography.bib}
\end{document}